\documentclass[conference]{IEEEtran}
\usepackage{amsmath,amsfonts}
\usepackage{algorithmic}
\usepackage{algorithm}
\usepackage{array}
\usepackage[caption=false,font=normalsize,labelfont=sf,textfont=sf]{subfig}
\usepackage{textcomp}
\usepackage{stfloats}
\usepackage{url}
\usepackage{verbatim}
\usepackage{graphicx}
\usepackage{cite}
\usepackage{multirow} 
\usepackage{balance}

\usepackage{array}
\newcommand{\PreserveBackslash}[1]{\let\temp=\\#1\let\\=\temp}
\newcolumntype{C}[1]{>{\PreserveBackslash\centering}p{#1}}
\newcolumntype{R}[1]{>{\PreserveBackslash\raggedleft}p{#1}}
\newcolumntype{L}[1]{>{\PreserveBackslash\raggedright}p{#1}}

\usepackage{amsmath}

\DeclareMathOperator*{\argmax}{arg\,max}

\begin{document}

\title{DQ4FairIM: Fairness-aware Influence Maximization using Deep Reinforcement Learning}

\author{
\IEEEauthorblockN{Akrati Saxena}
\IEEEauthorblockA{%
LIACS, Leiden University, \\The Netherlands \\
a.saxena@liacs.leidenuniv.nl}

\and

\IEEEauthorblockN{Harshith	Kumar Yadav}
\IEEEauthorblockA{%
IIT Ropar, India \\
e.23csz0002@iitrpr.ac.in}

\and

\IEEEauthorblockN{Bart Rutten}
\IEEEauthorblockA{%
Eindhoven University of Technology,\\ The Netherlands}

\and

\IEEEauthorblockN{Shashi Shekhar Jha}
\IEEEauthorblockA{%
IIT Ropar, India \\
shashi@iitrpr.ac.in}
}

\maketitle

\begin{abstract}
The Influence Maximization (IM) problem aims to select a set of seed nodes within a given budget to maximize the spread of influence in a social network. However, real-world social networks have several structural inequalities, such as dominant majority groups and underrepresented minority groups. If these inequalities are not considered while designing IM algorithms, the outcomes might be biased, disproportionately benefiting majority groups while marginalizing minorities. In this work, we address this gap by designing a fairness-aware IM method using Reinforcement Learning (RL) that ensures equitable influence outreach across all communities, regardless of protected attributes. Fairness is incorporated using a maximin fairness objective, which prioritizes improving the outreach of the least-influenced group, pushing the solution toward an equitable influence distribution. We propose a novel fairness-aware deep Reinforcement Learning (RL) method, called DQ4FairIM (Deep $Q$-learning for Fair Influence Maximization), that maximizes the expected number of influenced nodes by learning an RL policy. The learnt policy ensures that minority groups formulate the IM problem as a Markov Decision Process (MDP) and use deep Q-learning, combined with the Structure2Vec network embedding, earning together with Structure2Vec network embedding to solve the MDP. The main benefit of using RL for this problem is its generalization ability, wherein a policy learnt on a specific problem instance can be generalized to new problem instances without the need to learn the model from scratch. We perform extensive experiments on synthetic benchmark datasets and real-world networks to compare our method with fairness-agnostic and fairness-aware baselines. The results show that our method achieves a higher level of fairness while maintaining a better fairness-performance trade-off than baselines. Additionally, our approach learns effective seeding policies that generalize across problem instances without retraining, such as varying the network size or the number of seed nodes. 

\end{abstract}

\begin{IEEEkeywords}
Influence Maximization, Reinforcement Learning, Maximin Fairness. 
\end{IEEEkeywords}

\section{Introduction}

Influence Maximization (IM) is the problem of identifying a small subset of nodes in a social network that can achieve maximum spread of information, behavior, or influence through the network. This problem has several applications, including viral marketing \cite{nguyen2016stop, huang2019community}, public health campaigns (e.g., HIV prevention for homeless youth \cite{wilder2018end, yadav2018bridging}, general health awareness \cite{valente2007identifying}), financial inclusion \cite{banerjee2013diffusion}, political mobilization, and information dissemination \cite{jaouadi2024survey}. In critical applications, ensuring broad and equitable information reach is essential. However, due to resource constraints, it is infeasible to reach every individual at risk directly. The IM solutions, given a budget $k$, aim to select top influential nodes that can maximize the overall influence spread in the network \cite{jaouadi2024survey}.

Real-world social networks embed structural inequalities shaped by socioeconomic, demographic, or institutional factors that influence their evolution. Social theories such as homophily, preferential attachment, and the glass ceiling effect explain how these inequalities arise, leading to imbalanced access to social capital and visibility \cite{saxena2024fairsna}. As a result, such networks tend to be skewed, with majority groups occupying more central and well-connected positions, while minority groups remain peripheral and underrepresented. In \cite{saxena2024fairsna}, the authors discuss structural inequalities and biases present in social networks and their impact on network analysis algorithms, including IM. Network analysis algorithms that do not account for these inequalities frequently generate biased outcomes, specifically for minorities. 

Traditional IM approaches primarily optimize for overall influence without accounting for structural inequalities; therefore, small and marginalized groups, who benefit the most from attention and assistance, are frequently overlooked. Fairness-aware IM methods have emerged to address these disparities by ensuring that influence is more equitably distributed across different social groups \cite{saxena2024fairsna, stoica, farnad2020unifying}. In this work, we adopt a deep reinforcement learning (RL) approach to fair IM, leveraging the strengths of RL in sequential decision-making under uncertainty. Unlike heuristic or static methods, RL can dynamically learn policies that adapt to the underlying network structure and fairness objectives through interaction and feedback. While RL has shown success in network-based problems \cite{gajane2022survey}, such as link prediction \cite{nie2025local}, key node identification \cite{fan2020finding}, influence maximization \cite{manchanda2020gcomb,li2022piano}, community detection \cite{ni2025slrl}, and fake news mitigation \cite{xu2024harnessing}, it has not yet been explored for fair IM in networks with structural inequalities.

We propose DQ4FairIM (\textbf{Deep Q}-Learning for \textbf{Fair} \textbf{I}nfluence \textbf{M}aximization), a deep RL-based method designed to learn policies that select seed nodes to maximize overall influence and fairness. We model the problem as a Markov Decision Process (MDP), where the RL agent selects seed nodes sequentially until a budget $k$ is reached. The reward function combines marginal gains in influence and fairness, where fairness is defined using the maximin criterion that prioritizes the least-influenced community to promote equitable outreach. Community influence is measured as the ratio of influenced to total nodes, encouraging balanced spread across all groups. We use Deep $Q$-learning in combination with Structure2Vec network embeddings \cite{dai2016discriminative} to train the agent. The proposed method is evaluated on both synthetic benchmark datasets, including the Homophilic Barabási-Albert (HBA) model \cite{karimi} and the Obesity dataset \cite{wilder2018optimizing}, as well as real-world networks, such as Facebook \cite{leskovec2012learning} and Twitter \cite{macedo2024gender}.

A key advantage of using RL is its generalizability, where the model is trained on available networks and can be applied to previously unseen networks. Empirically, we observe that the RL agent effectively learns and adapts to underlying structural inequalities across different networks and performs well on unforeseen networks. Our method consistently outperforms both fairness-oblivious and fairness-aware baselines in terms of fairness, while achieving comparable outreach, often the highest or second-highest. Under higher influence probabilities, it surpasses all baselines in both outreach and fairness. Furthermore, we evaluate the model on evolving networks by training on smaller graphs and testing on progressively growing networks. Unlike other methods that require re-execution at each network snapshot, our approach maintains high performance without retraining. These results highlight the potential of RL for developing scalable and fair solutions to a broader class of network analysis problems.

\section{Related Work}\label{sec_rw}

In this section, we discuss fairness-agnostic and fairness-aware influence maximization methods.

\subsection{Fairness-agnostic IM}
 
In social network analysis, the influence maximization problem has been studied extensively. The IM problem is NP-hard \cite{kempe}, and under the Independent Cascade diffusion model, evaluating the influence of a seed set is \#P-hard \cite{chen2010scalable}. Kempe et al. \cite{kempe} introduced a greedy algorithm achieving a $(1 - 1/e - \varepsilon)$ approximation, but it was not scalable. To address this, Leskovec et al. \cite{leskovec2007cost} proposed CELF, a lazy-forward optimization that leverages submodularity to improve efficiency, which is 700 times faster than the standard greedy approach. IM algorithms have been broadly categorized into: (i) simulation-based methods using Monte Carlo techniques \cite{zhou2015upper, jiang2011simulated}, (ii) proxy-based approaches approximating influence functions \cite{liu2014influence, cheng2014imrank}, and (iii) sketch-based methods that are scalable and maintain theoretical guarantees \cite{tang2014influence, nguyen2016stop, ohsaka2014fast}. In practice, centrality measures \cite{saxena2020centralitysurvey} such as degree, PageRank, closeness, and k-shell \cite{saxena2018k} are also effectively used for influence estimation. 

Recently, reinforcement learning (RL) has emerged as a promising approach for network optimization problems, such as network diffusion \cite{fan2020finding}, IM \cite{khalil2017learning, manchanda2020gcomb}, and influence blocking maximization \cite{he2022reinforcement}. Khalil et al. \cite{khalil2017learning} introduced S2V-DQN, combining Structure2Vec embeddings with Deep Q-Learning to learn greedy solutions for network analysis problems. The proposed method uses the structure2vec approach to embed the network into low-dimensional node representations, and then adopts a deep reinforcement learning technique to train a Deep Q-Network (DQN). Finally, the top-k seed nodes can be chosen based on the seed scores provided by the DQN. 
Manchanda et al. \cite{manchanda2020gcomb} extended this method and proposed GCOMB, which uses a GCN to prune unpromising nodes and learn the low-dimensional embedding of good nodes. Next, they apply RL to identify seed nodes, demonstrating scalability to large real-world networks. Other RL-based approaches include EDRL-IM \cite{ma2022influence}, which integrates evolutionary algorithms and deep Q-learning to optimize the spread of influence.

Li et al. \cite{li2022piano} proposed the PIANO method, which incorporates network embedding and RL techniques to estimate the expected influence of nodes. They also showed that the models learned on similar types of networks generalize well, and a directly trained model can be used on a new, similar kind of network.
The PIANO method provides both retrainable and pretrained models (PIANO-S,  PIANO@$<d>$, and PIANO-E) for static and evolving networks. It outperforms state-of-the-art methods in terms of both efficiency and total influence spread. 
Chen et al. \cite{chen2021contingency} proposed contingency-aware IM, where node activation is probabilistic as each selected node has some uncertainty about the willingness of influence propagation. However, they assume uniform activation probabilities for all nodes, which is a main limitation in diverse settings. 
None of these RL-based methods considers the structural inequalities present in social networks and incorporates fairness into their objectives.

\subsection{Fairness-aware IM}
Fairness-agnostic IM methods, which focus solely on maximizing total outreach, often favor major communities and marginalize minorities \cite{saxena2024fairsna}. Stoica et al. \cite{stoica} introduced fairness-aware IM by formalizing two notions: fairness in early adopters (equal representation among seed nodes) and fairness in outreach (equal influence reach across communities). They further proposed three seeding strategies, called agnostic, parity, and diversity, and theoretically proved that if the budget is less than a specific threshold, the diversity seeding achieves a larger expected outreach than the agnostic seeding and gets close to the parity outreach \cite{stoica2020seeding}. 
Farnadi et al. \cite{farnad2020unifying} proposed diversity and maximin fairness criteria, aiming to balance influence proportionally to community sizes and to maximize the influence of the least-influenced group. They modeled fairness via mixed integer programming, incorporating fairness constraints directly into optimization. Tsang et al. \cite{tsang} demonstrated that fairness-aware IM is non-submodular and proposed a multi-objective approach that achieves near-optimal results with minimal trade-offs in outreach. Becker et al. \cite{Becker} proposed randomized seeding (using probabilistic strategies instead of deterministic strategies) in order to achieve a higher level of fairness. Fairness-aware network embedding methods, including adversarial network embedding \cite{khajehnejad2020adversarial}, FairWalk \cite{Rahman2019}, and CrossWalk \cite{khajehnejad2022crosswalk}, have also been used for fair IM. These methods apply clustering techniques, such as k-medoids, to select seed nodes in a group-balanced manner on the generated fair embeddings.

Some existing methods focus on the performance-fairness trade-off, which can be determined based on the application requirements. Anwar et al. \cite{anwar2021balanced} proposed a greedy approach that maximizes the objective function balancing outreach and fairness. Rahmattalabi et al. \cite{rahmattalabi2020fair} proposed a family of welfare functions with an inequity aversion parameter that can be varied to get the required trade-off between fairness and maximum outreach. They also observed that high welfare could be obtained without a significant reduction in the total outreach, as observed in \cite{tsang}. Teng et al. \cite{teng2020influencing} proposed disparity seeding, which focuses on both factors, maximizing the outreach and influencing the required fraction of the target community. In disparity seeding, the influential users are ranked using PageRank, Target HI-index, and Embedding index, and the seed nodes are selected using simulation-based learning. Context-aware fair IM has also been explored, including time-critical influence maximization \cite{ali2019fairness} and community-specific outreach targets \cite{teng2021influencing}. 

While most prior work focuses on group fairness, Fish et al. \cite{fish2019gaps} addressed individual fairness by maximizing the minimum probability of information access. They proposed greedy and heuristic algorithms under a maximin welfare objective function. In \textbf{greedy} solution, the probabilities are computed for each node, and the node that maximizes the objective function is added to the solution. However, the greedy solution is slower as the probabilities will be computed for each node, and therefore, the authors further proposed a fast heuristic solution using the distance function.

In contrast, our work introduces DQ4FairIM, a fairness-aware deep reinforcement learning framework for influence maximization that explicitly accounts for structural inequalities. By directly integrating group fairness into the reward function, our approach enables explicit control over the fairness–outreach trade-off while ensuring scalability and generalization across diverse network types.

\section{Preliminaries}\label{sec_pm}
This section outlines the prerequisites for understanding the problem formulation and the proposed solution. 

\subsection{Influence Propagation}\label{icmodel}

The diffusion process for influence propagation is simulated using the Independent Cascade (IC) model \cite{shakarian2015independent}, a widely used and well-established approach in this type of study. In this model, given a graph $G(V, E)$, each edge $(u, v)$ has an influence probability $p_{uv} \in [0,1]$ representing the likelihood that node $u$ activates node $v$. Nodes can be in either an active or inactive state. The influence spreads in discrete steps. Initially, a set of seed nodes $S$ is activated (e.g., they adopt a product or share a message). At each time step, newly activated nodes attempt to activate each of their inactive neighbors with the corresponding edge probability. Each node has only one chance to influence its neighbors. The process continues until no more nodes are activated. The outreach, denoted by $\sigma(G, S)$, is defined as the ratio of activated (influenced) nodes to the total number of nodes in the graph, given a seed set $S$.

In our work, we use the uniform IC model, where all edges have the same influence probability.

\subsection{Network Embedding}

Network embedding is a low-dimensional vector representation of the nodes of a graph. Formally, network embedding is a mapping $\phi: V \rightarrow \mathbb{R}^{d}$, where each node $v \in V$ is represented by a $d$-dimensional vector $\phi(v)$ in $\mathbb{R}^{d}$.

We use the Structure2vec embedding method \cite{dai2016discriminative}, a Graph Neural Network(GNN) approach for representing graph-structured data by generating vector representations for nodes in a graph. The key idea is to embed each node into a learned feature space by repeatedly aggregating information from its neighbours. Structure2Vec begins by initializing each node’s embedding and then repeatedly updates it by combining the node’s own features with the embeddings of its neighbouring nodes. This allows the model to capture both local structure and latent dependencies while learning embeddings that are optimized for the downstream prediction task. Specifically, we adopt the Structure2vec DE-MF (Discriminative Embedding using Mean Field) variant\cite{dai2017learning}. In the next section, we describe how these embeddings are used to define the $Q$-function in our RL framework.

\subsection{Markov Decision Process}

\textbf{Markov Decision Process (MDP)} is a mathematical framework to model the RL problem. It is formally defined as a tuple $(\mathcal{S}, \mathcal{A}, \mathcal{P}, \mathcal{R}, \gamma)$, where: $\mathcal{S}$ is the state space, $\mathcal{A}$ is the action space, $\mathcal{P}: \mathcal{S} \times \mathcal{A} \to \mathcal{S}$ is the state transition probability function, which takes a state and an action as input and returns the next state, $\mathcal{R}: \mathcal{S} \times \mathcal{A} \to \mathbb{R}$ is the reward function, which assigns a reward to each state-action pair, and $\gamma \in [0,1)$ is the discount factor. We consider a finite MDP that models a discrete-time stochastic control process, where at each time step the system is in some state, on which the agent takes an action, and the system transitions to a new state according to the transition function.

The solution to the MDP is a \textbf{policy} $\pi: \mathcal{S} \to \mathcal{A}$ which is a mapping from states to actions, indicating which action to take given a state. The \textbf{value function} $V^\pi$ represents the expected return that the agent receives when starting from state $s$ and following policy $\pi$ thereafter: $V^{\pi}(s) = \mathbb{E}_{\pi}\left[ G_t \mid S_t = s \right].$ 
The \textbf{action-value function} (or Q-function) $Q^\pi$ is the expected return when starting from state $s$, taking action $a$, and then following policy $\pi$:
\begin{equation}
Q^{\pi}(s, a) = \mathbb{E}_{\pi}\left[ G_t \mid S_t = s, A_t = a \right].  
\end{equation}

The main goal of reinforcement learning is to learn an optimal policy $\pi^*$ that maximizes the expected total reward over time.

\section{Problem Formulation}\label{problemformulation}
Let's consider a social network represented by a graph $G = (V, E)$, where $V$ is the set of nodes and $E$ is the set of edges. The network is partitioned into $C = \{C_1, C_2, \cdots, C_i, \cdots, C_l\}$ disjoint communities (also called groups) based on the given protected attribute, with each node belonging to exactly one community. Given a seed set $S \subseteq V$, let $\sigma_{C_i}(G, S)$ denote the fraction of influenced nodes in community $C_i$ after running the diffusion process from $S$. To quantify fairness in influence spread, we use the \textit{maximin fairness} criterion, defined as the minimum fraction of influenced nodes across all communities. It is computed as:

\begin{equation} \label{maxminfairness}
f(G, S) = \min_{\forall C_i \in C} \sigma_{C_i}(G,S)
\end{equation}

Maximin fairness ensures that the least-influenced community receives more benefits from the newly selected seed node, promoting equitable information outreach across all groups.

\vspace{2mm}

\noindent\textbf{Problem Statement.} Given a network $G = (V, E)$ and a budget $k$, the objective is to select a seed set $S \subseteq V$ with $|S| = k$ that maximizes the overall influence spread $\sigma(G, S)$, while ensuring maximal fairness in influence distribution across communities, measured using the maximin fairness criterion $f(G, S)$. Formally, the optimization problem is defined as:

\begin{subequations} \label{eq:optimization}
\begin{align}
    \max_{S} \quad & \sigma(G,S) \label{eq:optimization_a}\\
    \text{subject to} \quad & S \in \arg\max_{S'} f(G,S') \label{eq:optimization_b}\\
    & |S'| = k,\; S' \subseteq V \label{eq:optimization_c}
\end{align}
\end{subequations}

During the seed selection process, if multiple seed sets yield the same maximum fairness, any of those sets leading to maximum outreach can be chosen. Next, we model the fair IM problem using RL.

\subsection{Formulation using Reinforcement Learning (RL)}\label{formulationusingRL}

The formulation of IM as an MDP is explained below.

\textbf{Time step:} Each round of seed selection is modeled as a discrete time step $t$. In a single-round setting, one node is selected per round, and in a multi-round setting, $B$ nodes can be selected per round. The time horizon is indicated by $t = 1, \dots, T$ with $T = k$, where $k$ is the maximum number of seed nodes allowed. Thus, the agent selects one node at each time step until the seed set reaches size $k$. 

\textbf{State:} At time step $t$, the state $S_t$ is represented by the tuple $S_t = (G, C, X_t)$, where $G = (V, E)$ is a graph sampled from the pool of graphs $\mathcal{G}$, $C$ is the set of communities in the network, and $X_t \in {0,1}^{|V|}$ is a binary vector indicating which nodes have been selected as seed nodes up to time $t$, i.e., $X_t^v = 1$ if node $v$ is selected, and $0$ otherwise. Initially, no node is selected, and therefore, $X_1^v = 0$ for all $v \in V$. For implementation, $X_t$ may be extended to incorporate community membership or node attributes, e.g., $X_t \in {0,1}^{|V| + C}$ where $X_t^{v,i+1} = 1$ indicates that node $v$ belongs to community $i$.

\textbf{Action:} At each time step $t$, the agent selects a node to add to the seed set. The action is represented by a one-hot vector $a_t \in {0,1}^{|V|}$ with one non-zero entry indicating the chosen node (i.e., $\sum_{v=1}^{|V|} a_t^v = 1$). The action space is constrained by the current state, since an agent can only choose from nodes that have not been selected yet, i.e., $X_t^v = 0$. 

\textbf{State-action-transition probability}: The state transition is deterministic. When a node is selected at time step $t$, it is added to the seed set and the state is updated accordingly: $$X_{t+1} = X_t + a_t, \quad t=1,\cdots, T.$$

\textbf{Reward:} The total reward at the end of an episode is defined as the total influence outreach $\sigma(G, S)$ in the network $G$ resulting from the selected seed set $S$. However, this formulation implies that the agent would only receive a reward at the terminal time step $T$, and not during intermediate steps. This sparsity of rewards can significantly hinder the agent’s learning efficiency. To address this, we define an immediate reward at each time step using the marginal influence of the selected node. Specifically, the immediate reward is defined as: 

\begin{equation} r(G,X_t,a_t) = \sigma(G,S \cup \{v\}) - \sigma(G,S), \end{equation} 
where $v$ is the node selected at time $t$ ($v \in V$ and $a_t^v = 1$), and $S = \{u \in V \mid X_t^u = 1\}$ is the current seed set.

Here, $\sigma(\cdot)$ denotes the expected influence outreach. One practical challenge is computing $\sigma(\cdot)$, as it typically requires running a large number of Monte Carlo simulations (e.g., $m=1,000$), which is computationally expensive to do at every time step across all episodes. Instead, we estimate $\sigma(\cdot)$ using a small number of simulations. Although this introduces stochasticity, it remains suitable as the agent can learn an approximate reward function through repeated interactions.  
 
\begin{figure*}[t]
    \centering
    \includegraphics[width = \textwidth]{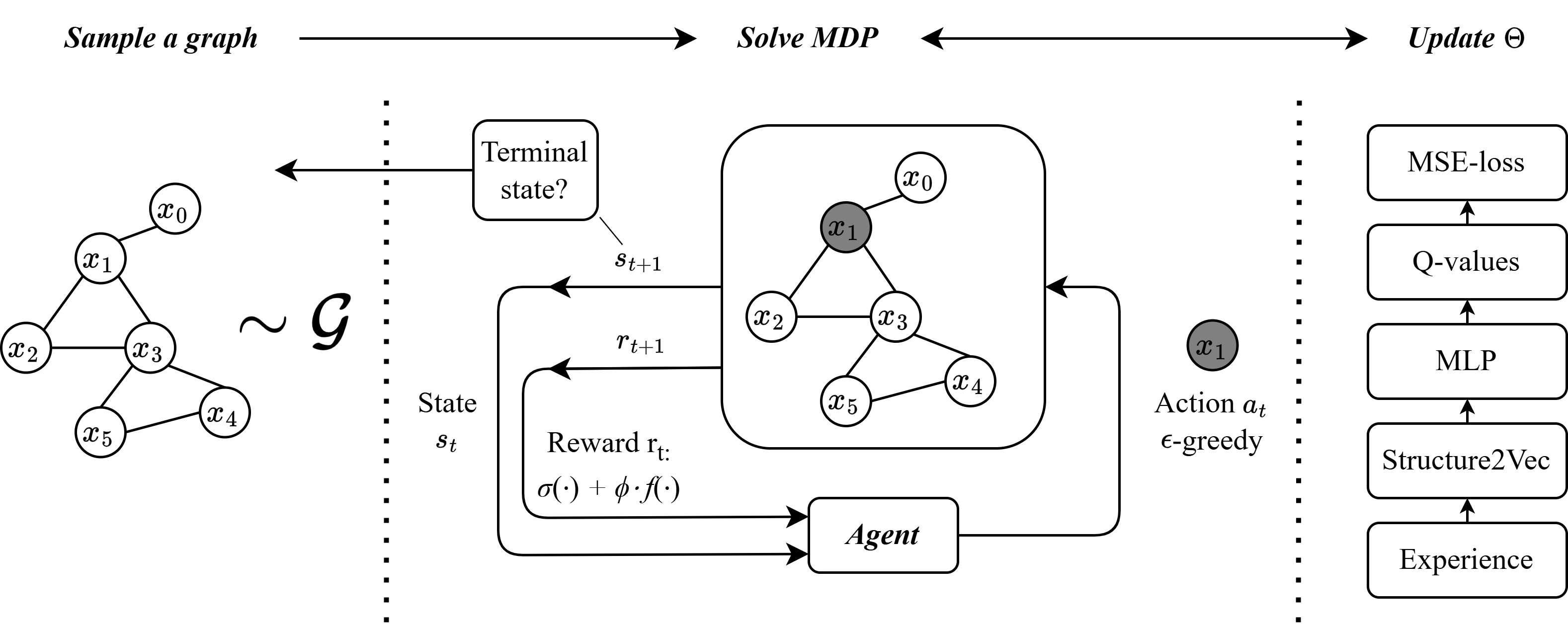}
    \caption{Graphical overview of the DQ4FairIM algorithm as described in Algorithm \ref{DQ4FairIM}. The process starts by randomly selecting a graph from the pool of graphs. This selected graph creates a new environment, and the agent then interacts with the environment and solves the MDP defined in Section \ref{formulationusingRL} for this specific graph. It chooses a new node (action) based on the $\epsilon$-greedy policy: it either selects a random node or a node with the highest $Q$-value. The reward it receives is based on both the expected influence outreach and the fairness measure. It picks a new graph at random once the terminal state is reached ($k$ seed nodes are selected) and a new episode begins. Along the way, the parameters of the neural network are updated with the samples stored in the Experience Replay Memory according to the mean squared error loss. First, the current state will be parameterized to an embedding space using the Structure2Vec mechanism. These embeddings will then be fed into a Multi-Layer Perceptron (MLP) to estimate the corresponding $Q$-values.}
    \label{fig: DQ4FairIMgraphic}
    \vspace{5mm}
\end{figure*}

\textbf{Enhancing fairness:} The formulation described above models the IM problem without considering fairness. To incorporate fairness, the agent should receive a reward not only based on the achieved influence but also on the fairness of its seed selection. Therefore, we define a combined reward function that includes both influence outreach and fairness, where fairness $f(G, S)$ is measured using the maximin criterion (refer to Eq. \ref{maxminfairness}). The total reward at the end of an episode is defined as:

\begin{equation} \label{eq: reward fairness} 
R(G, S) = \sigma(G, S) + \phi \cdot f(G, S), 
\end{equation} 
where $\phi \geq 0$ is a weighting parameter that controls the relative importance assigned to fairness. A higher value of $\phi$ increases the emphasis on fairness in the agent’s learning objective. We also redefine the immediate reward at each step $t$ to include marginal gains in both influence outreach and fairness: 

\begin{align}
r(G, X_t, a_t) &= \sigma(G, S \cup v) - \sigma(G, S)  \nonumber\\
&\quad + \phi \left( f(G, S \cup v) - f(G, S) \right) ,
\end{align}

\section{The Proposed Solution: DQ4FairIM} \label{section:DQ4FairIM} 
We propose a deep $Q$-learning algorithm, called \textbf{DQ4FairIM}, to solve the fair IM problem. We integrate deep $Q$-networks with structure2vec network embedding to solve the MDP formulation described in Section \ref{problemformulation}. While our algorithm follows the general principles of standard deep $Q$-learning, such as using a deep neural network to estimate the $Q$-values in combination with experience replay, it differs substantially in its neural network architecture as the model jointly computes network embeddings and estimates state-action values. 

Each node $v \in V$ is associated with a $d$-dimensional embedding vector $\mu_v$, which is recursively computed based on the graph structure and group membership in $C$. The initial input feature $x_v$ for each node includes its selection status $X_t$ (as defined in Section~\ref{formulationusingRL}). We use a variant of structure2vec where embeddings are initialized to zero, i.e., $\mu_v^{(0)} \in \mathbb{R}^d = \mathbf{0}$, and updated synchronously at each iteration for all nodes using the following rule: 

\begin{equation}
    \mu_v^{(t+1)} \leftarrow F\left(x_v, \{\mu_u^{(t)}\}_{u \in \mathcal{N}(v)}; \Theta\right),
\end{equation} 
where $\mathcal{N}(v)$ denotes the set of neighbors of node $v$ and $F$ is a nonlinear mapping, such as a kernel function or a neural network. After $T$ iterations, each node’s embedding $\mu_v^{(T)}$ captures information about its $T$-hop neighborhood. Specifically, we define the update function $F$ as: 

\begin{equation}
    \mu_v^{(t+1)} \leftarrow \text{relu}\left(\theta_1 x_v + \theta_2 \sum_{u \in \mathcal{N}(v)} \mu_u^{(t)}\right),
\end{equation}
where $\theta_1 \in \mathbb{R}^{(1+d) \times p}$ and $\theta_2 \in \mathbb{R}^{p \times p}$ are trainable model parameters. 

To compute the $Q$-value for a state-action pair $(S, a)$, we use the pooled graph embedding $\sum_{u \in V} \mu_u^{(T)}$ as the state representation and the node embedding $\mu_a^{(T)}$ as the action representation. The estimated $Q$-value is computed as:

\begin{equation}
    \hat{Q}(S, a; \Theta) = \theta_3^\top \text{ReLU}\left( 
    \left[ \theta_4 \sum_{u \in V} \mu_u^{(T)} \, \| \, \theta_5 \mu_a^{(T)} \right] 
    \right),
\end{equation}
where $\|$ denotes concatenation, $\theta_3 \in \mathbb{R}^{2d}$, and $\theta_4, \theta_5 \in \mathbb{R}^{d \times d}$ are trainable parameters.

This architecture enables DQ4FairIM to effectively learn policies that are both high-performing and fairness-aware in large graph-structured environments. The DQ4FairIM algorithm is presented in Algorithm~\ref{DQ4FairIM}. \textit{SimFairIC} function executes the IC model and computes the fair reward based on outreach and maximin fairness. For notational clarity, we represent $X_t$ in the algorithm as the set of selected seed nodes at time $t$, rather than a binary indicator vector.

While DQ4FairIM follows the general deep $Q$-learning framework~\cite{dai2017learning}, it incorporates additional features to solve the IM problem. First, we use the epsilon decay technique (in line 16 of Algorithm \ref{DQ4FairIM}) to balance the trade-off between exploration and exploitation. A high $\epsilon$ encourages exploration, which is particularly beneficial in early training stages when the agent has limited knowledge of the environment. As training progresses, it becomes advantageous for the agent to exploit its knowledge by selecting actions with higher estimated $Q$-values. The $\epsilon$ is gradually reduced by a decay factor $\eta$ after each episode, allowing the agent to increasingly exploit its learned policy. This decay continues until a predefined minimum value $\epsilon_{\text{min}}$ is reached. 
A second optimization is introduced in Line~11, where we update the neural network weights only at every $K$ steps instead of at each time step. This reduces computational overhead and improves training efficiency, as frequent updates can be costly without yielding significant improvements in performance.

An overview of all parameters used in DQ4FairIM is provided in Table \ref{tab:dq4fairimpara}. A graphical overview of the DQ4FairIM architecture, including a description of its components, is shown in Figure \ref{fig: DQ4FairIMgraphic}. 

\textbf{Time complexity of DQ4FairIM.} Given that the parameters $\Theta$ are updated in every step ($K=1$), the (training) time complexity of DQ4FairIM for a pool of graphs is $O(kE(b\cdot|V|\cdot |\Theta| + m\cdot |E|))$, where $k$: the budget, $E$: the number of episodes, $b$: the batch size, $|V|$: the number of nodes in a graph, $|E|$: the number of edges in a graph, $|\Theta|$: the number of weights for $\hat{Q}()$, and $m$: the number of Monte-Carlo simulations for the IC model. 

\begin{algorithm}[t]
\caption{DQ4FairIM: Deep $Q$-learning for fair IM}
\label{DQ4FairIM}
    \begin{algorithmic}[1]
\STATE Initialize $\mathcal{M}$, $\Theta$, $E$, $\epsilon$
\FOR{episode $e=1$ till $E$}
\STATE Draw a random graph $G$ from pool of graphs $\mathcal{G}$, set $R_0=0$
\STATE Initialize state $S_0 = (G,C,X_0)$, with seed set $X_0=\{\}$ 
\FOR{step $t=1$ till budget $k$}
\STATE{\footnotesize$ a_t =
\begin{cases}
\text{random node } v \in V\setminus S_t, & \text{with probability } \epsilon \\
\argmax_{v \in V \setminus S_t} Q(S_t, a; \Theta), & \text{with probability } 1-\epsilon\\
\end{cases}$}
\STATE Add node $a_t$ to solution: $X_{t+1}:= X_t \cup \{a_t\}$, $S_{t+1}=(G,C, X_{t+1})$
\STATE Calculate $R_t=\textsc{SimFairIC}(G,X_{t+1},C,\phi,f,m)$
\STATE Reward is marginal gain: $r_t= R_t-R_{t-1}$
\STATE Add tuple $(S_t, a_t, r_t, S_{t+1})$ to $\mathcal{M}$
\IF{$(e\cdot k +t) \mod K = 0$}
\STATE Sample random batch of transitions $B$ of size $b$ from $\mathcal{M}$
\STATE {\small Set $y_j =
\begin{cases}
r_j, & \hspace{-6mm} \text{for terminal } S_{j+1} \\
r_j + \gamma \max_{a} Q(S_{j+1},a; \Theta), & \text{otherwise}\\
\end{cases}$}
\\ \quad \quad \quad \quad \quad \quad \quad \quad \quad \textit{\# execute line 13  $\forall j \in B$}
\STATE Update $\Theta$ using Stochastic Gradient Descent over $(y_j - \hat{Q}(S_j,a_j; \Theta))^2$ for $B$
\ENDIF
\STATE Update exploration parameter: $\epsilon \leftarrow \max{(\eta\cdot\epsilon, \epsilon_{\text{min}})}$
\ENDFOR
\ENDFOR
\STATE return $\Theta$
\end{algorithmic}
\end{algorithm}

\begin{table}[t]
    \centering
    \caption{Overview of notations used in DQ4FairIM.}
    \label{tab:dq4fairimpara}
    \begin{tabular}{p{.85 cm}|p{6.95 cm}} \hline
    \textbf{Notation} & \textbf{Description} \\ \hline
    $\mathcal{G}$     &  Pool of graphs to train the RL model.\\ 
    $k$     & Budget size, i.e., the number of seed nodes to be selected \\
    $p_{uv}$     & Propagation probability of node $u$ to node $v$ in IC model. \\
    $\phi$ & Weighting parameter for fairness in reward function 
    \\
    $E$ & The number of episodes for training the RL agent. \\
    $\gamma$ & Discount parameter in the RL framework. 
    \\
    $\epsilon$ & The probability of selecting a random node as the next action according to the $\epsilon$-greedy policy. \\
    $\eta_\epsilon$  & Cooling down parameter of $\epsilon$. \\
    $\epsilon_\text{min}$ & The lowest value for $\epsilon$ 
    \\
    $d$ & Dimension of network embeddings \\
    $\Theta$ & The set of parameters to estimate the $Q$-function. \\
    $\mathcal{M}$ & Replay memory size \\
    $b$ & Batch size \\
    $\alpha$ & Learning rate for stochastic gradient descent \\
    $m$ & The number of (Monte-Carlo) simulations for the IC model \\
    $K$ & Every $K$ steps, the parameters of the function approximator are updated. \\ \hline
    \end{tabular}
\end{table}

\section{Experimental Setup}\label{experimentalsetup}
All experiments are performed on a system having the configuration of 2 x Intel Xeon Gold 6142 (2.60GHz/16-core/22MB/150W) with a DDR4 RAM of 60GB and a GPU of 6xGTX 1080Ti, 12GB (NVIDIA CUDA® Cores: 3584). 

The hyperparameters in RL model are set as: discount factor $\gamma=1$, initial exploration rate $\epsilon=1$ with decay rate $\eta_\epsilon=0.995$ down to a minimum $\epsilon_{\min}=0.05$, embedding size $d=64$, replay memory size $\mathcal{M}=2000$, batch size of $b=32$, learning rate $\alpha=0.001$ and $\phi=1$. The influence probability of the IC model is $0.1$ and $|S|=30$, if not otherwise mentioned. 

\textbf{Evaluation metrics.} The performance is evaluated using \textbf{Outreach}, i.e., defined as the fraction of influenced nodes in the network. \textbf{Fairness} is measured using the maximin fairness criterion (defined in Eq. \ref{maxminfairness}), which denotes the minimum fraction of influenced nodes across all communities. We execute 1000 Monte Carlo simulations from a given seed set to compute the average and standard deviation of Outreach and Fairness. Next, we discuss the datasets and baselines used in the experiments.

\subsection{Datasets}

The experiments are performed on synthetic (Homophily-BA and Obesity Prevention) and real-world networks (Facebook) as described below.

\textbf{Homophily-BA Networks:} We generate synthetic networks using the Homophily-BA model~\cite{karimi}, which produces homophilic scale-free networks containing majority and minority communities of specified proportions. In our experiments, nodes are divided in a 20:80 minority-to-majority ratio, with a homophily factor of $0.8$ and a preferential attachment parameter $0.2$. We construct 60 networks of 1,000 and 10,000 nodes, with an average degree of 8, and refer to these datasets as \textbf{HBA1k} and \textbf{HBA10k}, respectively. Each dataset has 60 networks, where 50 are used for training and 10 for testing. 

\textbf{Obesity dataset:} This dataset, introduced by Wilder et al. \cite{wilder2018optimizing}, simulates networks to model California's Antelope Valley, capturing real-world social network properties. It consists of 24 graphs, each with 500 nodes, where nodes are assigned attributes such as region, ethnicity, age, gender, and health status. Nodes with similar attributes are connected with a high probability, reflecting homophily \cite{saxena2025homophily}. We compare fairness based on gender, having two values, `male' and `female'.

\textbf{Real-world Networks:} To evaluate our model on real-world data, we use the \textbf{Facebook} ~\cite{leskovec2012learning} and \textbf{Twitter} (collected for Soccer communication in Portuguese) ~\cite{macedo2024gender} networks. In both datasets, gender (values: `male' or `female') is used as the sensitive attribute. For experimentation, we uniformly sample 10 network instances with 1,000 nodes each from both datasets, using six instances for training and four for testing.

\subsection{Baselines}

We compare our method with the following fairness-agnostic (1-4) and fairness-aware (5-8) baselines.

\begin{enumerate}
    \item CELF (Cost-Effective Lazy Forward) algorithm \cite{leskovec2007cost}: This is an optimized version of the greedy algorithm for IM. 
    \item Degree: Selects the top-$k$ nodes having the highest degree.
    \item Pagerank: Selects top-k nodes based on the Pagerank centrality.
    \item PIANO \cite{li2022piano}: PIANO incorporates network embedding and RL techniques to estimate the expected influence of nodes. In our work, we use the PIANO-S model, where the model is trained on all training graphs and tested on the testing graph set. 
     
    \item Parity \cite{stoica2020seeding}: Parity seeding is a fairness-driven strategy that modifies the selection thresholds for different groups to balance representation in the seed set. Overall, it selects nodes based on degree, ensuring that the group proportions of the seed set match those of the entire population. 
    \item Fair Pagerank: The parity-based seeding is applied on PageRank centrality instead of degree, ensuring that seed nodes are fairly selected from all communities. 

    \item Crosswalk \cite{khajehnejad2022crosswalk}: Crosswalk generates fairness-aware network embeddings, and selects the most centrally located nodes as seed nodes using the $k$-medoids clustering method.
    \item CEA \cite{ma2024fair}: This is a community-based evolutionary algorithm for fair influence maximization that uses a community-aware node selection strategy combined with tailored evolutionary operators (initialization, crossover, mutation) to select seed nodes. 

\end{enumerate}

\section{Results}

We trained \textbf{DQ4FairIM} on a pool of training graphs and \textit{evaluated its performance on unseen testing graphs}. During testing, the agent constructs a solution by iteratively selecting nodes with the highest predicted $Q$-value using the learned function $\hat{Q}(s, a, \Theta)$. All non-RL baseline methods are only executed on the testing graphs. Table~\ref{tab:outreach_fairness_all} presents the average outreach and maximin fairness scores for all methods; the highest values are in bold and the second highest are in italics. The standard deviation ranges between $0.0001$ and $0.0006$ across all cases. 

DQ4FairIM with $\phi = 1$ achieves the highest fairness across all datasets, including both synthetic and real-world networks. The outreach performance is comparable to the best-performing methods. In most cases, total outreach is maximized by CEA, Crosswalk, or DQ4FairIM ($\phi = 1$). 
We also evaluated the fairness of all methods using the disparity in addition to the maximin fairness metric. The disparity is computed as: 

\begin{equation*}
    Disparity=\max_{i,j \in \{1, 2, ... , |C|\}}\left | \sigma_{C_i}(G,S) - \sigma_{C_j}(G,S)  \right |
\end{equation*}

Table~\ref{tab:disparity} reports the disparity values across different datasets. The results show that DQ4FairIM ($\phi = 1$) consistently achieves the lowest (or second lowest) disparity among all methods, across both synthetic and real-world networks. This demonstrates that DQ4FairIM ($\phi = 1$) performs on par with baseline methods in terms of both maximin and disparity fairness constraints. 

\begin{table*}[t]
\caption{Average outreach and maximin fairness on synthetic and real-world networks.}
\label{tab:outreach_fairness_all}

\begin{tabular}{p{22mm}|p{10mm}p{10mm}|p{10mm}p{10mm}|p{10mm}p{10mm}|p{10mm}p{10mm}|p{10mm}p{10mm}}
\hline
\multicolumn{1}{c|}{\multirow{1}{*}{}} & \multicolumn{6}{c|}{\textbf{Synthetic}} & \multicolumn{4}{c}{\textbf{Real-world}}      \\ \hline
\multicolumn{1}{c|}{}                        & \multicolumn{2}{c|}{\textbf{HBA1k}}                                                                                                   & \multicolumn{2}{c|}{\textbf{HBA10k}}                                                                                                  & \multicolumn{2}{c|}{\textbf{Obesity}}                                                                                                 & \multicolumn{2}{c|}{\textbf{Facebook}}                                                                                                & \multicolumn{2}{c}{\textbf{Twitter}}                                                                                                 \\ \hline
\multicolumn{1}{c|}{Method}                        & \begin{tabular}[c]{@{}l@{}}Outreach\end{tabular} & \begin{tabular}[c]{@{}l@{}}Fairness\end{tabular} & \begin{tabular}[c]{@{}l@{}}Outreach\end{tabular} & \begin{tabular}[c]{@{}l@{}}Fairness\end{tabular} & \begin{tabular}[c]{@{}l@{}}Outreach\end{tabular} & \begin{tabular}[c]{@{}l@{}}Fairness\end{tabular} & \begin{tabular}[c]{@{}l@{}}Outreach\end{tabular} & \begin{tabular}[c]{@{}l@{}}Fairness\end{tabular} & \begin{tabular}[c]{@{}l@{}}Outreach\end{tabular} & \begin{tabular}[c]{@{}l@{}}Fairness\end{tabular} \\ \hline
CELF                                         &   0.1857                                                           &   0.1832                                                         &   0.1043                                                          &    0.0926                                                         &    0.0442                                                       &    0.0356                                                       &    0.8479                                                         &    0.8343                                                         &   \textit{0.1871}                                                       &   \textit{0.1508}                                                          \\
Degree                                      &    0.1914                                                          &    0.1618                                                          &   0.1102                                                           &   0.0976                                                           &    0.1158                                                          &    0.1063                                                          &   0.8414                                                           &   0.8263                                                           &   0.1734                                                           &   0.1294                                                           \\
Pagerank                                    &    0.1822                                                          &    0.1745                                                          &    0.0975                                                          &   0.0829                                                           &    0.1105                                                          &     0.1011                                                         &    0.8457                                                          &    0.8316                                                          &    0.1742                                                          &     0.1345                                                         \\
Parity                                      &    0.1917                                                          &    0.1829                                                        &     0.1097                                                          &    \textit{0.1007}                                                          &    0.1151                                                         &     \textit{0.1113}                                                      &   0.8413                                                           &   0.8264                                                           &    0.1769                                                          &   0.1364                                                           \\
Fair Pagerank                               &    0.1842                                                          &     0.1788                                                         &    \textit{0.1104}                                                          &    0.0975                                                          &     0.1103                                                         &     0.1071                                                         &     0.8459                                                         &     0.8317                                                         &    0.1822                                                          &     0.1415                                                         \\
CrossWalk                                   &    0.1859                                                          &     0.1691                                                         &     0.1091                                                         &    0.1003                                                        &     0.0899                                                         &     0.0793                                                         &     \textbf{0.8523}                                                &     \textit{0.8381}                                                &    0.1822                                                          &     0.1457                                                         \\
CEA                                         &    \textit{0.1971}                                                         &     \textit{0.1837}                                                         &   \textbf{0.1119}                                                  &    0.0999                                                          &    \textbf{0.1182}                                                 &    0.1075                                                          &    0.8451                                                          &   0.8308                                                           &   \textbf{0.1927}                                                  &    0.1457                                                          \\
PIANO                             &    0.1779                                                          &     0.1707                                                         &      0.1062                                                        &    0.0975                                                          &    0.1140                                                          &    0.1027                                                          &  \textit{0.8520}                                                            &  0.8378                                                            &    0.1802                                                          &    0.1470                                                         \\
DQ4FairIM                            & \textbf{0.2098}                                                    & \textbf{0.2010}                                                    &  0.1102                                                            &  \textbf{0.1010}                                                   & \textit{0.1169}                                                             & \textbf{0.1146}                                                         & \textbf{0.8523}                                                             &    \textbf{0.8382}                                                          &   0.1859                                                           &   \textbf{0.1561} \\ \hline                                                          
\end{tabular}
\vspace{4mm}
\end{table*}

\begin{table}[]
\caption{Disparity fairness on synthetic and real-world networks.}
\label{tab:disparity}
\begin{tabular}{p{16mm}|p{9mm}|p{9.5mm}|p{9mm}|p{9.5mm}|p{9mm}}
\hline
\multicolumn{1}{c|}{} & \multicolumn{3}{c|}{\textbf{Synthetic}} & \multicolumn{2}{c}{\textbf{Real-world}} \\ \hline
\multicolumn{1}{c|}{} & \textbf{HBA1k} & \textbf{HBA10k} & \textbf{Obesity} & \textbf{Facebook} & \textbf{Twitter} \\ \hline

CELF &0.0200 & 0.0155 &0.0170  & \textbf{0.0375} & 0.0456  \\
Degree & 0.0226 & 0.0167 & 0.0185 & 0.0417 & 0.0555 \\
Pagerank & 0.0232 & 0.0194  & 0.0181 & 0.0505 & 0.0513 \\
Parity & \textit{0.0137} & 0.0120 & 0.0076 & 0.0471 & 0.0569 \\
Fair Pagerank & 0.0221 & 0.0142 & \textit{0.0064} & 0.0414 & 0.0500 \\
CrossWalk & \textit{0.0137} & \textit{0.0116} & 0.0213  & 0.0389 & 0.0459 \\
CEA & 0.0183 & 0.0160 & 0.0205& 0.0394 & 0.0508 \\
PIANO & 0.0241 & 0.0121 & 0.0222 & 0.0392 & \textit{0.0414} \\
DQ4FairIM  & \textbf{0.0137} & \textbf{0.0116}  & \textbf{0.0043} & \textit{0.0387} & \textbf{0.0375} \\ \hline
\end{tabular}
\vspace{4mm}
\end{table}

\begin{figure}[t]
    \flushleft
    \subfloat[Outreach]{%
        \includegraphics[width=0.4\textwidth]{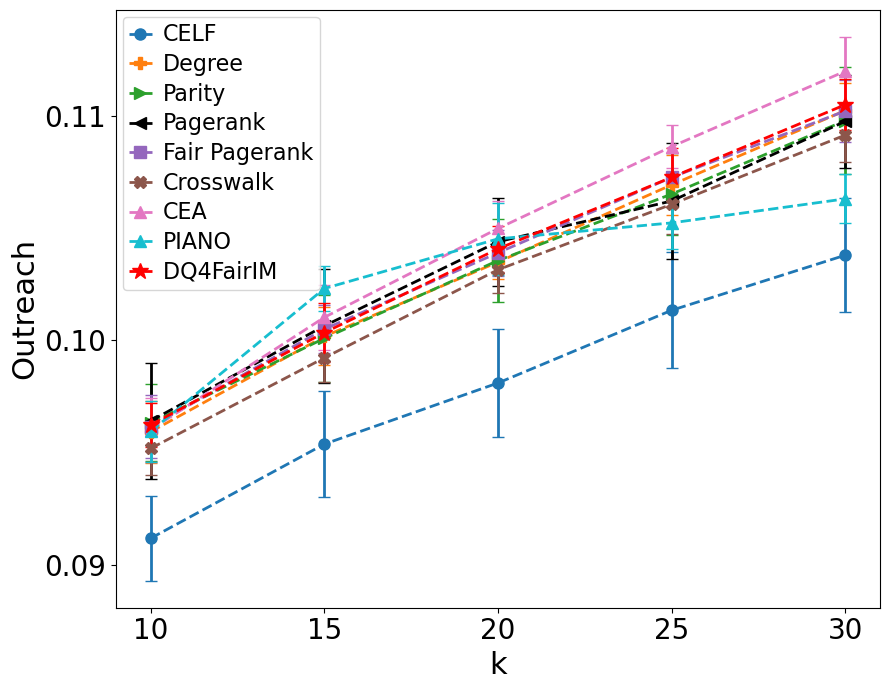}%
        \label{fig:outreach_vary_k}%
    }
    \hspace{0.5cm} 
    \subfloat[Fairness]{%
        \includegraphics[width=0.4\textwidth]{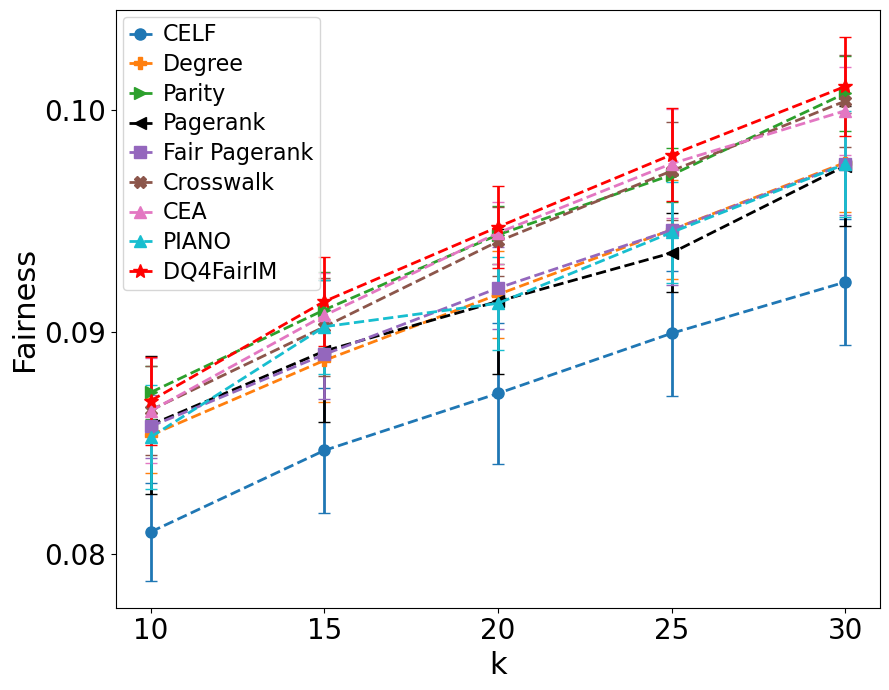}%
        \label{fig:fairness_vary_k}%
    }
    \caption{Outreach and fairness for varying seed nodes on the HBA10k dataset.}
    \label{fig:varyingk}
\end{figure}

\textbf{Varying the Number of Seed Nodes.} 
We evaluate how influence outreach and fairness vary with different seed set sizes ($k$). Results are shown in Figure \ref{fig:varyingk} for the HBA10k dataset. We observe that across all values of $k$, DQ4FairIM consistently achieves the highest fairness while maintaining influence outreach comparable to that of other methods. Furthermore, we observe that as $k$ increases, fairness-aware methods tend to outperform fairness-agnostic ones in terms of total outreach. This finding aligns with prior work on influence blocking \cite{saxena2023fairness}, which suggests that fairness-aware strategies not only ensure equitable outcomes but also enhance overall performance over time, acting as a catalyst for improved performance over time.

\textbf{Varying Influence Probability.} We further evaluate all methods under varying influence probabilities in the range $[0.1, 0.15, 0.25]$. The results for the HBA10k dataset are presented in Table~\ref{veryinginfprob}. The CELF method was not completed within 24 hours and is therefore excluded from the comparison. Across all influence probability settings, \textbf{DQ4FairIM} consistently achieves the highest fairness and outreach compared to baseline methods. We performed this experiment on other datasets and also for other influence probabilities $(0.05, 0.3)$ and received similar results. Notably, as the influence probability increases, the outreach achieved by the \textbf{DQ4FairIM} grows substantially and outperforms other methods by a significant margin.

\begin{table}[]
\caption{Outreach and fairness on HBA10k dataset for varying influence propagation probability in the IC model.}\label{veryinginfprob}
\setlength{\tabcolsep}{2pt} 
\renewcommand{\arraystretch}{1.1}

\begin{tabular}{p{16mm}|C{10mm}C{10mm}|C{10mm}C{10mm}|C{10mm}C{10mm}}
\hline
& \multicolumn{2}{c|}{Inf Prob = 0.15}
& \multicolumn{2}{c|}{Inf Prob = 0.20}
& \multicolumn{2}{c}{Inf Prob = 0.25} \\
Method & Out. & Fair. & Out. & Fair. & Out. & Fair. \\
\hline
CELF                                                      &  --                                                            &   --                                                           &  --                                                            &  --                                                            &   --                                                           &  --                                                            \\
Degree                                                    &0.3941                                                              & 0.3672                                                            & 0.6158                                                             &  0.5888                                                            & 0.7560                                                             &   0.7342                                                              \\
Parity                                                         &   0.3942                                                           &  0.3677                                                            & 0.6159                                                             &  0.5886                                                            &  0.7560                                                            &  0.7342                                                            \\
Pagerank                                                  
&  0.3941                                                            & 0.3669                                                             & 0.6158                                                             &  0.5886                                                            &  0.7560                                                            & 0.7342                                                             \\
Fair Pagerank                                             
&  0.3940                                                            & 0.3674                                                             & 0.6159                                                             & 0.5888                                                             &  0.7560                                                            &   0.7342                                                           \\
CrossWalk                                                
&  0.3947                                                            & 0.3684                                                             &  0.6159                                                            &  0.5888                                                            &  0.7560                                                            &  0.7342                                                            \\
CEA                                                       
&  0.3958                                                            &  0.3689                                                             &  0.6161                                                            &  0.5888                                                            &  0.7560                                                            &  0.7340                                                            \\

DQ4FairIM   &  \textbf{0.3997}                                                            & \textbf{0.3705}                                                             &  \textbf{0.6176}                                                            & \textbf{0.5901}                                                             & \textbf{0.7573}                                                             & \textbf{0.7343}   \\ \hline                                                         
\end{tabular}
 \vspace{2mm}
\end{table}

\begin{table*}[t]
\centering
\caption{\textbf{Generalizability results:} We train DQ4FairIM on the HBA10k dataset and evaluate it on larger evolving networks containing 20k–50k nodes, while all baseline methods (except PIANO) are executed on each new dataset for comparison.} 
\label{veryingsize}
\begin{tabular}{p{16mm}|C{10mm}C{10mm}|C{10mm}C{10mm}|C{10mm}C{10mm}|C{10mm}C{10mm}}
\hline
      & \multicolumn{2}{c|}{HBA20k}  & \multicolumn{2}{c|}{HBA30k}   & \multicolumn{2}{c}{HBA40k}    &
      \multicolumn{2}{c}{HBA50k} \\ 
Method & Out. & Fair. & Out. & Fair. & Out. & Fair. & Out. & Fair. \\ \hline
CELF & -- & -- & -- & -- & -- & --&--&-- \\
Degree & 0.1016 & 0.0899 & 0.0997 & 0.0882 & 0.0983 & 0.0868 & 0.0974 & 0.0861 \\
Parity & 0.1016 & 0.0917 & 0.0995 & 0.0895 & 0.0982 & 0.0877 & 0.0974 & 0.0867\\
Pagerank & 0.1022 & 0.0904 & 0.0999 & 0.0884 & 0.0986 & 0.0871 & 0.0976 & 0.0862 \\
Fair Pagerank & 0.1010 & 0.0905 & 0.0997 & 0.0889 & 0.0983 & 0.0879 & 0.0979 & 0.0871 \\
CrossWalk & 0.1016 & \textit{0.0918} & 0.0999 & \textit{0.0897} & 0.0986 & 0.0879 & 0.0976 & 0.0871\\
CEA & \textbf{0.1038} & 0.0912 &  \textbf{0.1015} & 0.0895 & \textbf{0.1002} & \textit{0.0880} & \textbf{0.0989}& 0.0870\\
PIANO & \textit{0.1024} & 0.0899 & \textit{0.1010} &0.0881  & \textit{0.0997} & 0.0869 & \textit{0.0982} & 0.0861\\
DQ4FairIM  & 0.1018 & \textbf{0.0920} & 0.0999 & \textbf{0.0898} & 0.0989 & \textbf{0.0882} & 0.0980 & \textbf{0.0888}  \\ \hline
\end{tabular}
\vspace{2mm}
\end{table*}

\begin{figure}[!ht]
    \flushleft
    \subfloat[Outreach]{%
        \includegraphics[width=0.4\textwidth]{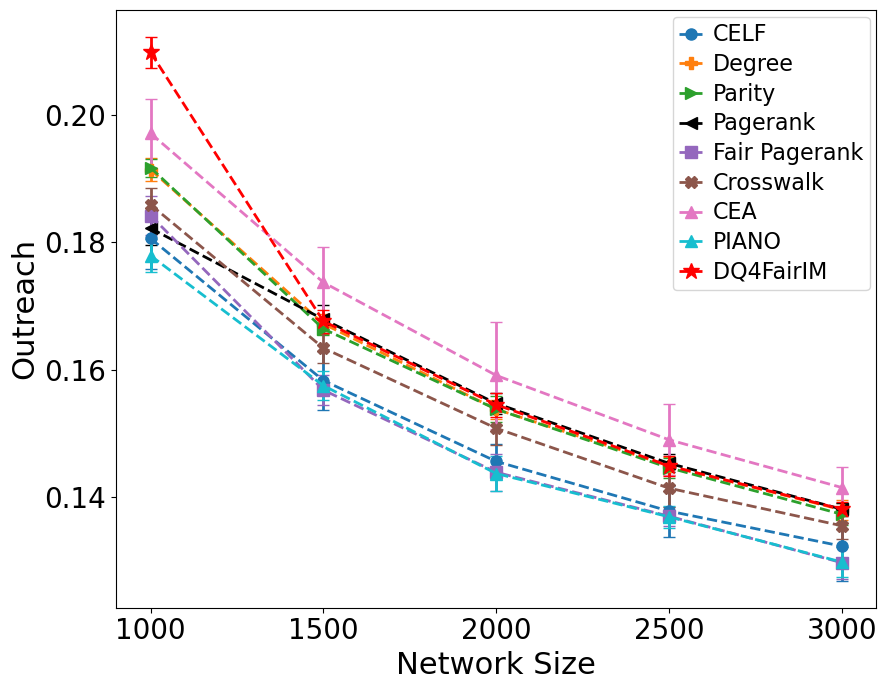}%
        \label{fig:outreach_vary_size}%
    }
    \hspace{0.5cm} 
    \subfloat[Fairness]{%
        \includegraphics[width=0.4\textwidth]{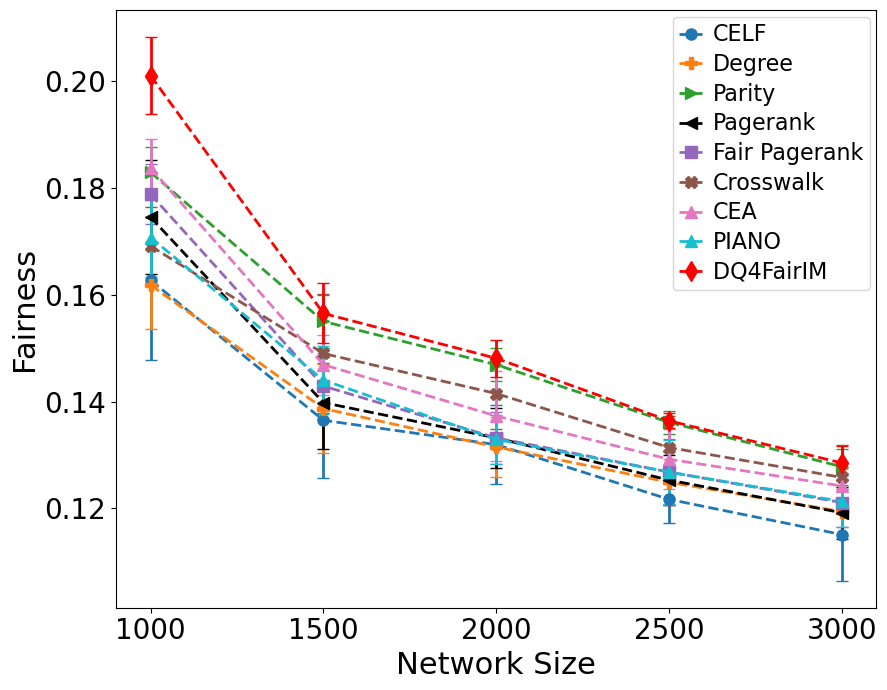}%
        \label{fig:fairness_vary_size}%
    }
    \caption{\textbf{Generalizability results:} Outreach and fairness on evolving graphs where the DQ4FairIM and PIANO models are trained on the HBA1k dataset and tested on bigger size networks, having 1000--3000 nodes. However, other baselines are executed on all new datasets to compute the outreach and fairness.} 
    \label{fig:varyinggraphsize}
\end{figure}

\begin{figure}[htbp]
    \centering

    \begin{minipage}{0.48\linewidth}
        \centering
        \subfloat[$\phi=0.75$]{%
            \includegraphics[width=\linewidth]{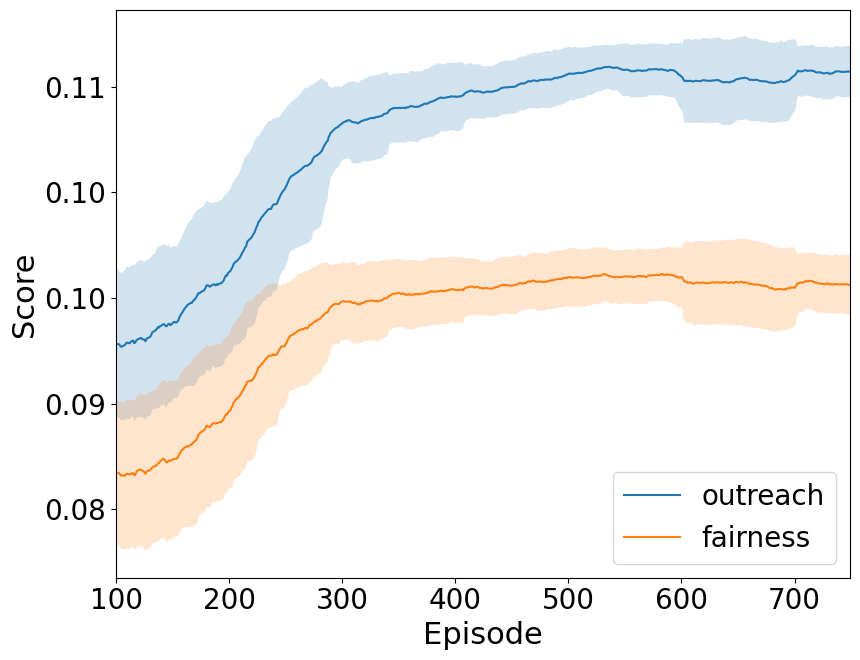}%
            \label{fig:avg_phi_075}%
        }
    \end{minipage}\hfill
    \begin{minipage}{0.48\linewidth}
        \centering
        \subfloat[$\phi=1$]{%
            \includegraphics[width=\linewidth]{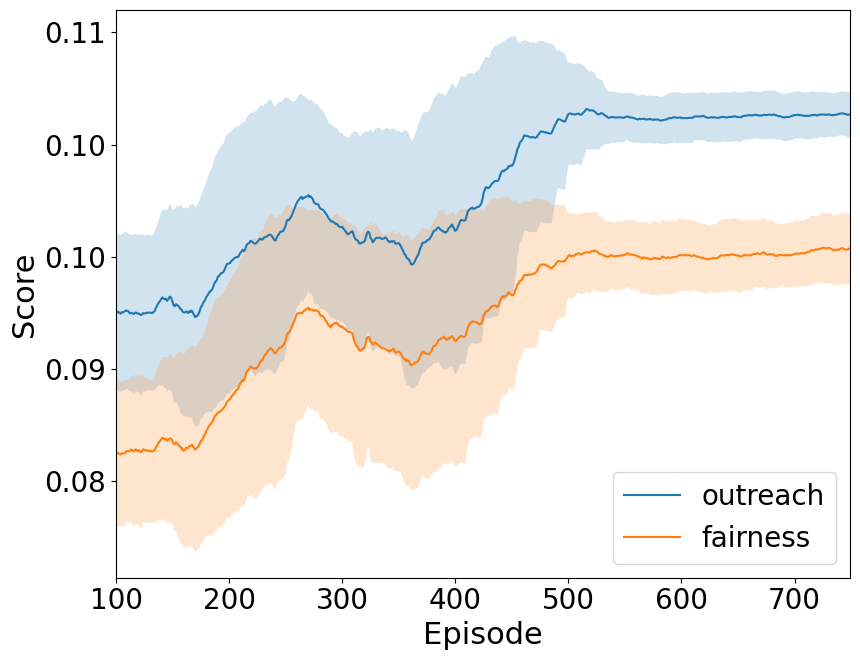}%
            \label{fig:avg_phi_1}%
        }
    \end{minipage}

    \vspace{4mm}

    \begin{minipage}{0.48\linewidth}
        \centering
        \subfloat[$\phi=0.75$]{%
            \includegraphics[width=\linewidth]{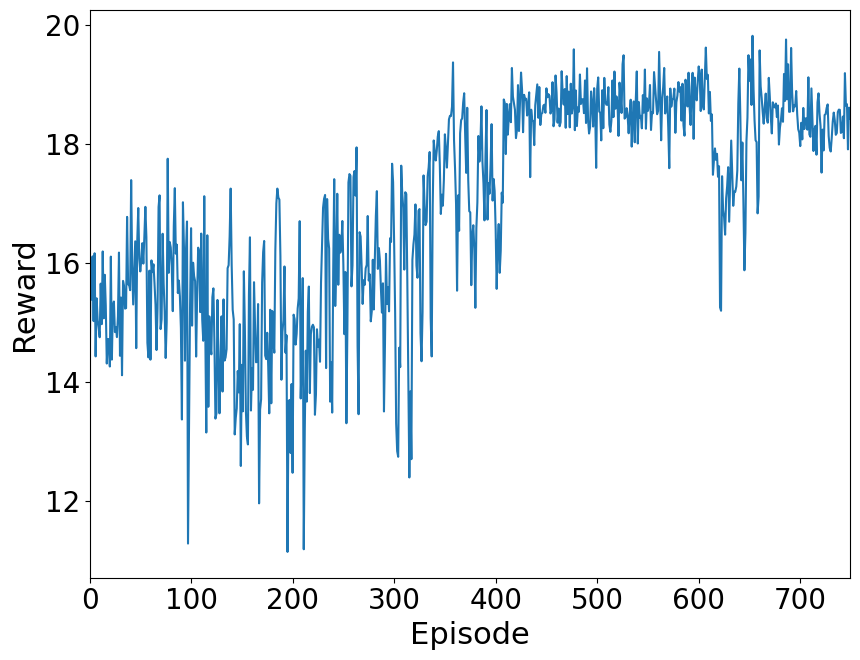}%
            \label{fig:reward_phi_075}%
        }
    \end{minipage}\hfill
    \begin{minipage}{0.48\linewidth}
        \centering
        \subfloat[$\phi=1$]{%
            \includegraphics[width=\linewidth]{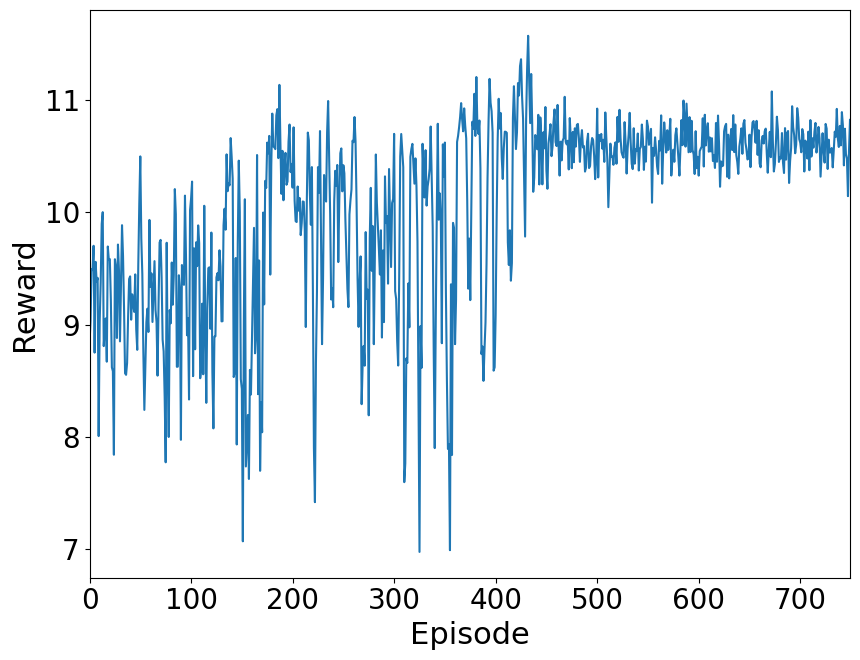}%
            \label{fig:reward_phi_1}%
        }
    \end{minipage}

    \caption{(a)--(b): Outreach and fairness, and (c)--(d): Reward while training DQ4FairIM for different levels of $\phi$ on the HBA10k dataset.}
    \label{fig:score_vs_episodes}
\end{figure}

\subsection{Generalization of DQ4FairIM}

\textbf{Performance on evolutionary networks.} 
Real-world networks are highly dynamic, often significantly grow in size over time. A key advantage of reinforcement learning (RL)-based approaches for IM is their ability to generalize to larger, unseen network instances without requiring retraining, unlike traditional algorithms that must be rerun for each network snapshot as the network evolves. To assess the scalability of our proposed method, we train DQ4FairIM on the HBA1k training dataset and use the learned $Q$-network to generate solutions for larger unseen test datasets of 1500, 2000, 2500, and 3000 nodes, having 10 networks each, which are drawn from the same HBA distribution. PIANO is also trained and tested in a similar manner, while all other baselines are executed on each new dataset separately. 
Figure~\ref{fig:varyinggraphsize} presents the results for both influence outreach and maximin fairness across these networks. While all methods, including CEA, Pagerank, Parity, and DQ4FairIM, achieve comparable outreach, DQ4FairIM consistently exceeds baseline methods in terms of fairness even on unseen evolving networks.

Similarly, to evaluate the generalizability on larger networks, we train DQ4FairIM and PIANO on the HBA10k training dataset and use the learned $Q$-network to generate solutions for network datasets having 10 networks of 20k, 30k, 40k nodes created using the HBA model with the same hyperparameters. Table~\ref{veryingsize} shows both the outreach and fairness across these networks. The CELF method was not completed within 24 hours and is therefore excluded from the comparison. Despite comparable outreach across all methods, DQ4FairIM clearly stands out in terms of fairness, outperforming all baselines. 

These results suggest that DQ4FairIM generalizes effectively from smaller graphs to much larger ones, assuming the graph structure remains similar. This highlights a practical advantage that fair IM solutions on large-scale networks can be efficiently derived by training on smaller, representative subsets. This not only significantly reduces computational overhead during training but also maintains strong performance in both fairness and total outreach.

\subsection{Ablation study}

To examine how the DQ4FairIM agent adapts under varying fairness constraints, we analyze its learning dynamics across different values of the fairness weight parameter $\phi \in {0, 0.25, 0.5, 0.75, 1}$. The model is trained for 750 episodes using the default hyperparameters as mentioned in Section \ref{experimentalsetup}. 
Figures~\ref{fig:score_vs_episodes}~(a-b) show the outreach and fairness over the course of training for $\phi =0.75$ and $\phi=1$. For better readability, we plot the 50-episode rolling mean with standard deviation for outreach and fairness. Figures \ref{fig:score_vs_episodes}~(c-d) show the total reward (outreach + $\phi \times$ fairness) per episode on training graphs. As expected in RL training, reward values are initially low due to exploratory behavior, but begin to rise sharply after approximately 300-400 episodes and eventually stabilize and converge.

Similar results are observed for other $\phi$ values and across different networks. The curves for non-zero $\phi$ ($\phi = 0.5$ to $1$) are closely aligned, suggesting that the agent effectively learns to balance fairness and outreach as long as fairness is assigned non-zero weight. Across all $\phi$ values, the agent shows a clear learning progression. Importantly, higher $\phi$ values consistently lead to increased fairness, without significantly compromising the overall influence outreach.

\section{Conclusion}
We propose DQ4FairIM, a novel deep reinforcement learning-based framework, for fair influence maximization, where the fairness is defined using the maximin fairness constraint that aims to maximize the influence spread in the minimally influenced group. By formulating the IM problem as a Markov Decision Process and incorporating fairness in the reward function, our method balances the trade-off between maximizing influence spread and ensuring fair outreach across different population groups. Experimental results on synthetic and real-world networks demonstrated that DQ4FairIM consistently finds fairer solutions compared to all fairness-agnostic and fairness-aware baselines, without compromising on overall influence spread. Moreover, the method generalizes well to larger and unseen graphs, indicating strong scalability and practical utility. The model’s capacity to learn from past networks and rapidly adapt to new ones makes it especially useful in dynamic domains, such as public health interventions and social awareness campaigns.

Future work can explore RL for fair influence maximization under different diffusion models, such as the Linear Threshold model \cite{kempe} or edge-weighted cascades \cite{saxena2015understanding}, and context-specific conditions, such as time-critical diffusion \cite{ali2019fairness} or target-based fair IM \cite{teng2020influencing}.

\balance

\bibliographystyle{IEEEtran}
\bibliography{mybib}

\end{document}